# LDW-SCSA: Logistic Dynamic Weight based Sine Cosine Search Algorithm for Numerical Functions Optimization

Türker TUNCER[1]

[1]Department of Digital Forensics Engineering, Technology Faculty, Firat University, Elazig, Turkey

turkertuncer@firat.edu.tr

*Abstract*—Particle swarm optimization (PSO) and Sine Cosine algorithm (SCA) have been widely used optimization methods but these methods have some disadvantages such as trapped local optimum point. In order to solve this problem and obtain more successful results than others, a novel logistic dynamic weight based sine cosine search algorithm (LDW-SCSA) is presented in this paper. In the LDW-SCSA method, logistic map is used as dynamic weight generator. Logistic map is one of the famous and widely used chaotic map in the literature. Search process of SCA is modified in the LDW-SCSA. To evaluate performance of the LDW-SCSA, the widely used numerical benchmark functions were utilized as test suite and other swarm optimization methods were used to obtain the comparison results. Superior performances of the LDW-SCSA are proved success of this method.

*Keywords*—Sine Cosine optimization; Logistic map; Dynamic weight generation; Swarm optimization; Numerical functions optimization.

## 1. Introduction

The name of the age we live in is the information age. In this age, information technologies are frequently used to solve real world problems. Some of the real world problems do not have a mathematical solution. To solve these, optimization methods have been used in the literature and real world applications. Optimization is the process of searching a global optima solution of a problem in a finite search space. Optimization algorithms consist of two sub-classes and these are gradient based optimization and meta-heuristic optimization algorithms. In the real-world applications, some problems cannot be solved by using mathematically approaches. In order to solve these problems, meta-heuristic optimization algorithms have been used. In the past two decade, researchers proposed many optimization algorithms and optimization has become hot-topic research area. Meta-heuristic algorithms do not require gradient knowledge and they call fitness (objective) function repeatedly in order to find global minimum. These algorithms try to narrow the search space and find an effective solution. The well-known swarm optimization algorithm is Particle Swarm Optimization (PSO). PSO is proposed by Kenedy et al. [1] in 1995 and this algorithm is nature-inspired heuristic optimization algorithm. In the PSO, social behavior of individuals of fish and bird swarms were mathematically modelled. Besides the PSO, many optimization algorithms such as artificial neural network [2,3], ant colony optimization (ACO) [4], moth-flame optimization (MFO) algorithm [5], artificial bee colony (ABC) algorithm[6] ,firefly algorithm [7], sine-cosine algorithm (SCA) [8], genetic algorithm (GA) [9], bat algorithm (BA) [10], differential evaluation (DE) [11], biogeography-based optimization (BBO) [12] , harmony search (HS) [13], gravitational search algorithm (GSA) [14], krill herd algorithm (KH) [15], etc., were proposed in the literature.

## 1.1. Contributions

The main aim of swarm optimization algorithms to calculate best solution but some of them trapped local best solutions. To solve this problem, chaotic maps have been used to calculate velocity of these algorithms [16]. CDW-PSO [16] clearly demonstrated that, dynamic weights were increased performance of the PSO. In this paper, chaotic dynamic weights are used in the SCA and logistic map is utilized as weight generator. The proposed method called as Logistic Dynamic Weights based Sine-Cosine Search Algorithm (LDW-SCSA). LDW-SCSA is a mathematical heuristic optimization method and the experiments clearly demonstrated that, this method is a successful meta-heuristic search method.

## 2. Preliminaries

In this section, the related methods of LDW-SCSA are mentioned.

## 2.1. PSO

PSO is the most used swarm optimization method in the literature and it has several variations. In this method, the movements of bird and fish swarms are modeled in order to find global optimum value. In the PSO, firstly particles are generated randomly in range of search space and all particles are updated until to reach global optimum value or maximum iteration. Eq. 1 and 2 describe mathematically particles updating of the PSO.

$$v_{id}^{t+1} = wv_{id}^t + c_1 rand_1(P_{id}^t - x_{id}^t) + c_2 rand_2(pbest_d^t - x_{id}^t) \qquad (1)$$

$$x_{id}^{t+1} = x_{id}^t + v_{id}^{t+1} \qquad (2)$$

Where $c_1$, $c_2$ are acceleration of the particles, $w$ is weight, $v$ is velocity, $rand_1$ and $rand_2$ are uniformly generated random numbers in range of [0,1], $p_{best}$ is position of pest particle, $x$ is position of particle, $i$ refers i[th] value, d is dimension index and t is time. Shi and Eberhart [17] proposed weighted PSO to increase success of the PSO and steps of this method are given below.

Step 1: Initialization. Generate particles randomly in range of lower and upper bound.

Step 2: Evaluate each particle by using objective function and calculate gbest and pbest.

Step 3: Update position of particles by using Eq. 1-2.

Step 4: Update gbest and pbest.

Step 5: Repeat steps 3-4 until the global optimum value or maximum iterations is reached.

**2.2. SCA**

Sine-cosine optimization method is a mathematic based swarm optimization and this method was proposed by Mirjalili in 2015 [18]. Mathematical description of SCA is given Eq. 3 [18-19].

$$x_{id}^{t+1} = \begin{cases} x_{id}^t + r_1 \sin(r_2) |r_3 pbest_d^t - x_{id}^t|, r_4 < 0.5 \\ x_{id}^t + r_1 \cos(r_2) |r_3 pbest_d^t - x_{id}^t|, r_4 \geq 0.5 \end{cases} \qquad (3)$$

In Eq. 3, $r_1$, $r_2$, $r_3$ and $r_4$ are randomly generated. $r_1$ determines direction of movements particles, $r_2$ determines how far to move inward or outward to reach the global optimum, $r_3$ determines stochastic weight and $r_4$ provides the transition between sine and cosine components.

Steps of SCA are given below.

Step 1: Initialization. Generate particles randomly in range of lower and upper bound.

Step 2: Evaluate each particle by using objective function and calculate pbest.

Step 3: Update position of particles by using Eq. 3.

Step 4: Update pbest.

Step 5: Update $r_1$, $r_2$, $r_3$ and $r_4$.

Step 6: Repeat steps 3-5 until the global optimum value is found or maximum iterations is reached.

## 2.3. Logistic Map

Chaos is one of the phenomena of nonlinear mathematics. In particular, chaotic maps and chaotic fractals are frequently used in information technologies. The computer sciences researchers generally use chaos in the information security and optimization algorithms. In the optimization algorithms, chaos provides many advantages because it provides uniform distributions. In the literature, many chaotic maps have been presented, one of the most known chaotic map is logistic map. Logistic map is a basic and effective chaotic map and Eq. 4 mathematically define logistic map [20-21].

$$t_{i+1} = rx_i(1 - t_i), r \in [3.57,4], x \in (0,1) \tag{4}$$

Seed values of this map are r and $t_1$ values. Random number are generated by using these values. In Eq. 4, r is chaos multiplier, $t_1$ is initial value and t is randomly generated sequence.

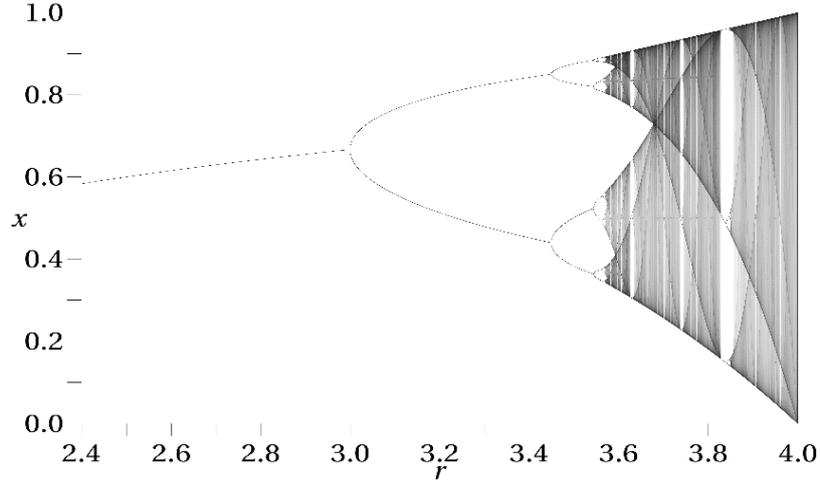

**Fig. 1.** Bifurcation diagram of logistic map [21].

## 3. The Proposed LDW-SCSA Method

In this paper, a novel chaotic weighted SCA is proposed and this method is a mathematical based metaheuristic search method. In the LDW-SCSA, weights are generated dynamically by using logistic map. The main aim of LDW-SCSA is to achieve more successful results than SCA. The LDW-SCSA consists of dynamic weight generation, weight based particle updating and optimal particle selection. The steps of LDW-SCSA are given below.

Step 1: Generate initial positions of the particles randomly.

$$p_i = rand[0,1] \times (UB - LB) + LB \tag{5}$$

Step 2: Evaluate all particles and select pbest.

Step 3: Calculate initial weight by using Eq. 6 and 7.

$$w_1 = eps \tag{6}$$

$$w_1 = \begin{cases} 1 - eps, w_1 = 1 \\ 0 + eps, w_1 = 0 \end{cases} \tag{7}$$

Where w(1) is initial weight and eps is 2.2204 x 10$^{-16}$. Eps is a constant ant it has widely used in the MATLAB for defining very small number.

Step 4: Generate weights by using logistic map.

$$w_{i+1} = 4w_i(1 - w_i) \tag{8}$$

Step 5: Generate $r_1$, $r_2$, $r_3$ and $r_4$ randomly. Range of $r_1$, $r_2$ and $r_3$ are [-2,2] and range of $r_4$ is [0,1].

Step 6: Calculate velocity of each particles by using Eq. 9.

$$v_i = \begin{cases} w_i r_1 \sin(r_2) |r_3 p_{best} - x_i|, r_4 < 0.5 \\ w_i r_1 \cos(r_2) |r_3 p_{best} - x_i|, r_4 \geq 0.5 \end{cases} \tag{9}$$

Step 7: Update position of particles by using Eq. 10.

$$p_i = p_i w_i + v_i + r_4 p_{best} w_i \tag{10}$$

Step 8: If particles exceed lower bound or upper bound, generate new particles in range of lower bound and upper bound randomly.

Step 9: Calculate fitness value of these particles and update pbest.

Step 10: Repeat steps 4 and 9 until the global optimum value is found or maximum iterations is reached.

r1, r2, r3, r4, w, LB and UB are utilized as parameters of LDW-SCA. LB and UB are determined range of search space and these are used in all of the swarm optimization methods. r1, r2, r3 and r4 have been generated randomly and w is generated by logistic map. Dynamically weight generation increases performance of SCA. LDW-SCA

```
1: Initialize parameters (LB is lower bound, UB is upper bound, MI is maximum iteration, pn is number
   of particles )
2: Randomly generate position of each particle in the swarm.
3: Evaluate each particles by using fitness function and find pbest in the swarm
4: w_1 = |pbest / (UB-LB)|
5: if w_1=1 then
6:     w_1=w_1 - eps
7: else if w_1=0 then
8:     w_1 = w_1 + eps
9: End if
10: for iter=1:MI do
11:     w_{iter+1} = 4w_{iter}(1 - w_{iter})
12:     for i=1:pn do
13:         Randomly generate r_1, r_2, r_3 and r_4
14:         if r_3<0.5 then
15:             v_i = w_{iter} r_1 sin(r_2) |r_3 pbest - p_i|
16:         else
17:             v_i = w_{iter} r_1 cos(r_2) |r_3 pbest - p_i|
18:         End if
19:         p_i = p_i w_{iter} + v_i + p_{best} w_{iter} r_4
20:         if p_i < LB or p_i > UB then
21:             Randomly assigned position to p_i in range of the problem
22:         End if
23:         Evaluate particle by using fitness function
24:         if fitness value of p_i better than pbest then
25:             p_i is pbest
26:         End if
27:         if fitness value of pbest is desired values then
28:             break;
29:         End if
30:     End for i
31: End for iter
```

Fig. 2. Pseudo code of the LDW-SCA

## 4. Experimental Results

In order to evaluate the performance of LDW-SCA, a test set were created which consisted of unimodal and multimodal numerical benchmark functions. This test set consists of 7 unimodal and 6 multimodal benchmark functions. The unimodal numerical functions used are given below [16,22].

1- Sphere function

$$f_1(x) = \sum_{i=1}^{n} x_i^2, x \in [-100,100]^n \qquad (11)$$

Sphere function is one of the widely used unimodal numerical functions in optimization. Global optimum of this function is $[0]^n$.

2- Schwefel 2.22 function

$$f_2(x) = \sum_{i=1}^{n} |x_i| + \prod_{i=1}^{n} |x_i|, x \in [-10,10]^n \qquad (12)$$

This function is one of the widely used function of schwefel's numerical function group. Global optimum of this function is $[0]^n$.

3- Schwefel 1.2. function

$$f_3(x) = \sum_{i=1}^{n} \left( \sum_{j=1}^{i} x_j \right)^2, x \in [-100,100]^n \qquad (13)$$

This function is one of the widely used function of schwefel's numerical function group. Global optimum of this function is $[0]^n$.

4- Schwefel 2.21 function

$$f_4(x) = \max\{|x_i|, 1 \leq i \leq n\}, x \in [-100,100]^n \qquad (14)$$

This function is one of the widely used function of schwefel's numerical function group. Global optimum of this function is $[0]^n$.

5- Rosenbrock's function

$$f_5(x) = \sum_{i=1}^{n-1} [100(x_{i+1} - x_i^2)^2 + (x_i - 1)^2], x \in [-30,30]^n \qquad (15)$$

Global optimum of this function is $[0]^n$.

6- Step function

$$f_6(x) = \sum_{i=1}^{n} |x_i + 0.5|^2, x \in [-100,100]^n \qquad (16)$$

Global optimum of this function is $[-0.5]^n$.

7- Noise function

$$f_7(x) = \sum_{i=1}^{n} ix_i + rand(0,1), x \in [-1.28,1.28]^n \qquad (17)$$

Global optimum of this function is $[0]^n$.

Multimodal numerical functions are given below [16,22].

1- Rastrigin's function

$$f_8(x) = \sum_{i=1}^{n} (x_i^2 + 10\cos(2\pi x_i) + 10), x \in [-5.12, 5.12]^n \qquad (18)$$

Global optimum of this function is $[0]^n$.

2- Ackley's function

$$f_9(x) = -20 \exp\left(-0.2 \sqrt{\frac{1}{n} \sum_{i=1}^{n} x_i^2}\right) - \exp\left(\frac{1}{n} \sum_{i=1}^{n} \cos(2\pi x_i)\right) + 20 + e \qquad (19)$$

$x \in [-32,32]^n$

Global optimum of this function is $[0]^n$.

3- Griewank's function

$$f_{10}(x) = \frac{1}{4000} \sum_{i=1}^{n} x_i^2 - \prod_{i=1}^{n} \cos\left(\frac{x_i}{\sqrt{i}}\right), x \in [-600, 600]^n \qquad (20)$$

Global optimum of this function is $[0]^n$.

4- Generalized Penalized 1 function

$$f_{11}(x) = \frac{\pi}{n} \left\{ 10\sin(\pi y_1) + \sum_{i=1}^{n-1} (y_i - 1)^2 [1 + 10\sin^2(\pi y_{i+1})] + (y_n - 1)^2 \right\} + \qquad (21)$$

$$\sum_{i=1}^{n} u(x_i, 10, 100, 4)$$

$$y_i = 1 + \frac{1}{4}(x_i + 1), u(x_i, a, k, m) = \begin{cases} k(x_i - a)^m, x_i > a \\ k(-x_i - a)^m, x_i < -a, x \in [-50,50]^n \\ 0, -a \leq x_i \leq a \end{cases}$$

Global optimum of this function is $[0]^n$.

5- Generalized Penalized 2 function

$$f_{12}(x) = \frac{1}{10}\left\{\sin^2(3\pi x_1) + \sum_{i=1}^{n}(x_i - 1)^2[1 + \sin^2(3\pi x_i + 1)] + (x_n - 1)^2[1 + \sin^2(2\pi x_n)]\right\} + \sum_{i=1}^{n} u(x_i, 5, 100, 4) \quad (22)$$

Where $u(x_i, a, k, m) = \begin{cases} k(x_i - a)^m, x_i > a \\ k(-x_i - a)^m, x_i < -a, x \in [-50,50]^n \\ 0, -a \leq x_i \leq a \end{cases}$

Global optimum of this function is $[0]^n$.

6- Alphine function

$$f_{13}(x) = \sum_{i=1}^{n}\left(|x_i \sin(x_i) + \frac{1}{10}x_i|\right), x \in [-10,10]^n \quad (23)$$

Global optimum of this function is $[0]^n$.

To evaluate of the LDW-SCA, the defined 13 numerical benchmark functions have been used. Performance of LDW-SCSA with different iterations and particles are shown in Table 1. Performance of LDW-SCSA with variable particles and 500 iterations

| Fun | Criteria | Max Iteration (MI) and particles (P) | | | | | |
|---|---|---|---|---|---|---|---|
| | | MI=500 P=10 | MI=500 P=20 | MI=500 P=30 | MI=500 P=40 | MI=500 P=50 | MI=500 P=60 |
| $f_1$ | Mean | 0 | 0 | 0 | 0 | 0 | 0 |
| | SD | 0 | 0 | 0 | 0 | 0 | 0 |
| $f_2$ | Mean | 1.6862e-215 | 3.1529e-255 | 1.6989e-269 | 1.6243e-273 | 2.1022e-292 | 7.2629e-295 |
| | SD | 0 | 0 | 0 | 0 | 0 | 0 |
| $f_3$ | Mean | 0 | 0 | 0 | 0 | 0 | 0 |
| | SD | 0 | 0 | 0 | 0 | 0 | 0 |
| $f_4$ | Mean | 3.6617e-248 | 2.3478e-253 | 2.1956e-280 | 1.0073e-289 | 2.2187e-293 | 9.9873e-304 |
| | SD | 0 | 0 | 0 | 0 | 0 | 0 |
| $f_5$ | Mean | 0 | 0 | 0 | 0 | 0 | 0 |
| | SD | 0 | 0 | 0 | 0 | 0 | 0 |
| $f_6$ | Mean | 0.0173 | 0.0167 | 0.0033 | 3.5782e-04 | 2.7767e-04 | 2.4298e-04 |
| | SD | 0.0385 | 0.0450 | 0.0078 | 7.9898e-04 | 7.5846e-04 | 5.1264e-04 |
| $f_7$ | Mean | 2.0578e-04 | 9.3213e-05 | 7.1311e-05 | 3.7401e-05 | 2.9754e-05 | 2.4796e-05 |
| | SD | 2.0325e-04 | 8.7549e-05 | 5.4361e-05 | 3.2985e-05 | 2.3949e-05 | 1.5564e-05 |
| $f_8$ | Mean | 0 | 0 | 0 | 0 | 0 | 0 |
| | SD | 0 | 0 | 0 | 0 | 0 | 0 |
| $f_9$ | Mean | 8.8818e-16 | 8.8818e-16 | 8.8818e-16 | 8.8818e-16 | 8.8818e-16 | 8.8818e-16 |
| | SD | 0 | 0 | 0 | 0 | 0 | 0 |
| $f_{10}$ | Mean | 0 | 0 | 0 | 0 | 0 | 0 |
| | SD | 0 | 0 | 0 | 0 | 0 | 0 |
| $f_{11}$ | Mean | 0.5387 | 0.2485 | 0.1661 | 0.1623 | 0.1610 | 0.1518 |
| | SD | 1.1957 | 0.6650 | 0.5149 | 0.5711 | 0.5874 | 0.5702 |
| $f_{12}$ | Mean | 0.0108 | 0.0090 | 0.0078 | 0.0065 | 0.0030 | 0.0026 |
| | SD | 0.0161 | 0.0186 | 0.0150 | 0.0098 | 0.0080 | 0.0065 |
| $f_{13}$ | Mean | 5.5404e-237 | 2.8026e-262 | 4.6505e-271 | 2.7046e-276 | 1.0035e-291 | 1.0966e-296 |
| | SD | 0 | 0 | 0 | 0 | 0 | 0 |

In order to compare the proposed algorithm with the other widely used swarm optimization algorithms, these numerical benchmark functions are used which are listed in Table 1. Moth-flame optimization (MFO) [5] algorithm, artificial bee colony (ABC) algorithm [6], sine-cosine algorithm (SCA) [8], biogeography-based optimization (BBO) [12] and krill herd algorithm (KH) [15] are used to obtain comparisons. Population, maximum iteration, dim of each object and other parameters of these algorithms are listed in Table 2.

Table 2. Parameters of the algorithms.

| Algorithms | Population | Number of maximum iteration | Dim of each object | Other |
|---|---|---|---|---|
| MFO | 40 | 500 | 30 | t is random [-2,1] |
| ABC | 40 | 500 | 30 | Limit=200 |
| SCA | 40 | 500 | 30 | $r_1$, $r_2$, $r_3$ and $r_4$ randomly generated. |
| BBO | 40 | 500 | 30 | Mu=0.005, m=0.8 |
| KH | 40 | 500 | 30 | $N^{max}$=0.01 $V_f$=0.02 $D^{max}$=0.005 |
| LDW-SCA | 40 | 500 | 30 | $r_1$, $r_2$, $r_3$ and $r_4$ randomly generated. w is generated by logistic map |

These algorithms and the LDW-SCSA were applied on the numerical benchmark functions which are listed in Table 1. The comparisons showed in Table 3.

Table 3. Performance comparison results.

| f | Criteria | MFO | ABC | SCA | BBO | KH | LDW-SCA |
|---|---|---|---|---|---|---|---|
| $f_1$ | Mean | $2 \times 10^3$ | $3.15 \times 10^{-5}$ | $1.11 \times 10^1$ | $7.17 \times 10^0$ | $5.19 \times 10^{-1}$ | **0** |
|  | S.D. | $4.22 \times 10^3$ | $3.47 \times 10^{-5}$ | $1.83 \times 10^1$ | $4.79 \times 10^0$ | $2.76 \times 10^{-1}$ | **0** |
| $f_2$ | Mean | $3.51 \times 10^1$ | $4.42 \times 10^{-3}$ | $1.62 \times 10^{-2}$ | $6.82 \times 10^{-1}$ | $3.84 \times 10^0$ | **$1.62 \times 10^{-273}$** |
|  | S.D. | $2.27 \times 10^1$ | $1.40 \times 10^{-3}$ | $2.12 \times 10^{-2}$ | $2.28 \times 10^{-1}$ | $2.11 \times 10^0$ | **0** |
| $f_3$ | Mean | $1.86 \times 10^4$ | $1.78 \times 10^4$ | $2.10 \times 10^4$ | $6.85 \times 10^3$ | $3.31 \times 10^2$ | **0** |
|  | S.D. | $1.40 \times 10^4$ | $2.92 \times 10^3$ | $1.12 \times 10^4$ | $1.67 \times 10^3$ | $1.41\ 10^2$ | **0** |
| $f_4$ | Mean | $6.76 \times 10^1$ | $4.35 \times 10^1$ | $7.37 \times 10^1$ | $1.20 \times 10^0$ | $6.09 \times 10^0$ | **$1 \times 10^{-289}$** |
|  | S.D. | $9.49 \times 10^0$ | $5.05 \times 10^0$ | $2.11 \times 10^1$ | $4.50 \times 10^0$ | $1.46 \times 10^0$ | **0** |
| $f_5$ | Mean | $9.59 \times 10^2$ | $4.36 \times 10^1$ | $1.70 \times 10^6$ | $8.49 \times 10^2$ | $8.96 \times 10^0$ | **0** |
|  | S.D. | $1.04 \times 10^3$ | $4.41 \times 10^1$ | $4.86 \times 10^6$ | $5.24 \times 10^2$ | $1.08 \times 10^2$ | **0** |
| $f_6$ | Mean | $1.01 \times 10^3$ | **$5.82 \times 10^{-5}$** | $3.69 \times 10^1$ | $1.48 \times 10^1$ | $4.61 \times 10^{-1}$ | $3.57 \times 10^{-4}$ |
|  | S.D. | $3.19 \times 10^3$ | **$9.74 \times 10^{-5}$** | $8.84 \times 10^1$ | $2.59 \times 10^1$ | $1.56 \times 10^{-1}$ | $7.98 \times 10^{-4}$ |
| $f_7$ | Mean | $6.31 \times 10^0$ | $3.03 \times 10^{-1}$ | $6.65 \times 10^{-1}$ | $8.31 \times 10^{-2}$ | $7.76 \times 10^{-2}$ | **$3.74 \times 10^{-5}$** |
|  | S.D. | $8.85 \times 10^0$ | $8 \times 10^{-2}$ | $1.81 \times 10^0$ | $4.53 \times 10^{-2}$ | $3.17 \times 10^{-2}$ | **$3.29 \times 10^{-5}$** |
| $f_8$ | Mean | $1.59 \times 10^2$ | $5.09 \times 10^0$ | $5.91 \times 10^1$ | $3.21 \times 10^2$ | $1.45 \times 10^1$ | **0** |
|  | S.D. | $2.66 \times 10^1$ | $2.32 \times 10^0$ | $3.65 \times 10^1$ | $6.47 \times 10^0$ | $1.07 \times 10^1$ | **0** |
| $f_9$ | Mean | $8.86 \times 10^{-8}$ | $1.54 \times 10^{-1}$ | $1.15 \times 10^1$ | $1.69 \times 10^0$ | $5.52 \times 10^0$ | **$8.88 \times 10^{-16}$** |
| | S.D. | 6.38 x 10$^{-8}$ | 1.43 x 10$^{-1}$ | 9.95 x 10$^{0}$ | 4.70 x 10$^{-1}$ | 1.40 x 10$^{0}$ | **0** |
| f$_{10}$ | Mean | 8.59 x10$^{-1}$ | 2.60 x 10$^{-2}$ | 1.04 x 10$^{0}$ | 1.08 x 10$^{0}$ | 1.48 x 10$^{-1}$ | **0** |
| | S.D. | 1.44 x 10$^{-1}$ | 2.86 x 10$^{-2}$ | 2.94 x 10$^{-1}$ | 7.83 x 10$^{-2}$ | 4.04 x 10$^{-2}$ | **0** |
| f$_{11}$ | Mean | 4.69 x 10$^{0}$ | **1.93 x 10$^{-5}$** | 4.17 x 10$^{6}$ | 3.42 x 10$^{-1}$ | 3.41 x 10$^{0}$ | 1.62 x 10$^{-1}$ |
| | S.D. | 1.81 x10$^{0}$ | **4.33 x 10$^{-5}$** | 1.38 x 10$^{7}$ | 4.44 x 10$^{-1}$ | 1.20 x 10$^{0}$ | 5.71 x 10$^{-1}$ |
| f$_{12}$ | Mean | 1.10 x 10$^{1}$ | **1.01x 10$^{-5}$** | 1.51 x 10$^{7}$ | 6.52 x 10$^{-1}$ | 3.85 x 10$^{-2}$ | 6.50 x 10$^{-3}$ |
| | S.D. | 9.65 x10$^{0}$ | **1.04 x 10$^{-5}$** | 2.87 x 10$^{7}$ | 2.18 x 10$^{-1}$ | 1.41 x 10$^{-1}$ | 9.80 x 10$^{-3}$ |

In the Table 2 all the functions were run 10 times and the best values were recorded. The mean and standard deviation values obtained are shown in Table 3. LDW-SCSA has produced the best value in 8 functions. LDW-SCSA is the modified and improved form of SCA. Comparisons of LDW-SCSA and SCA are shown in the convergence graph. 30 particles were used to obtain these convergence graphs.

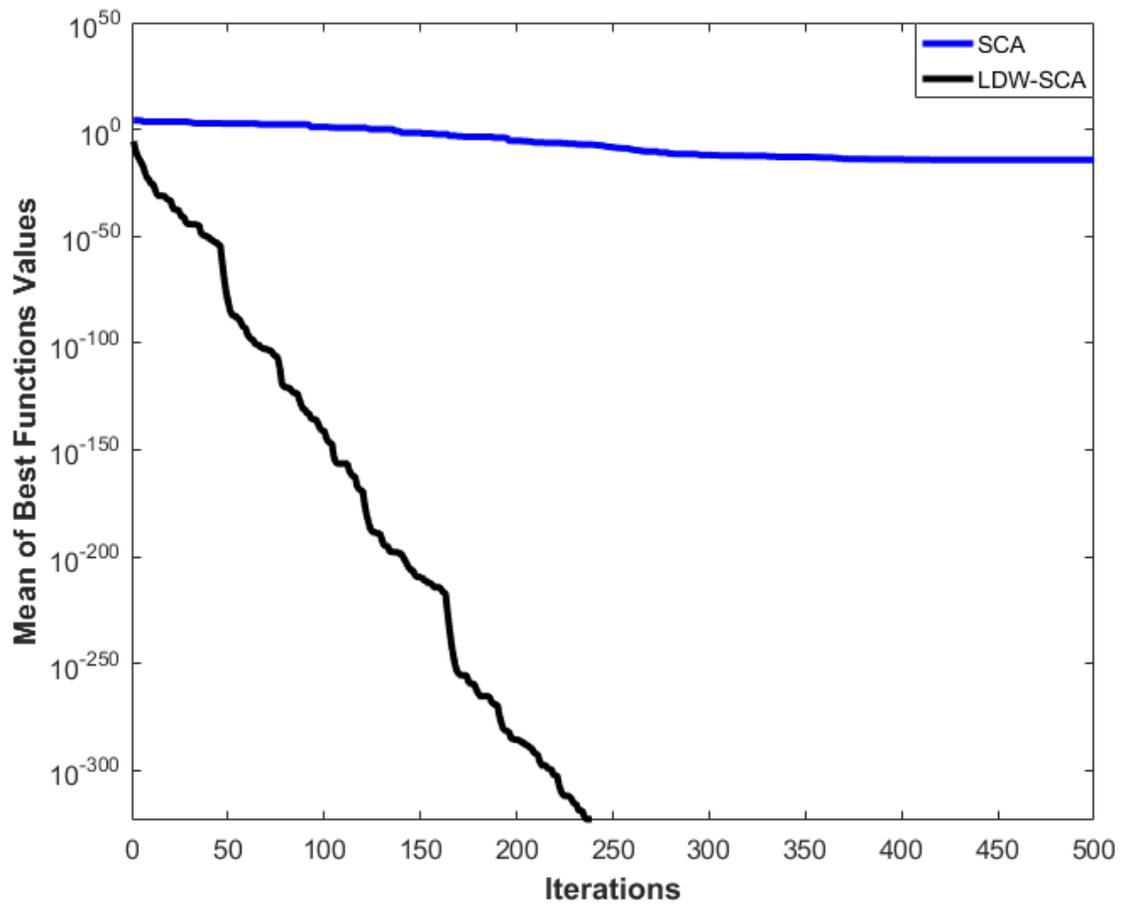

Fig. 3. Comparison of minimization performance of Sphere function ($f_1$) with Dim=30.

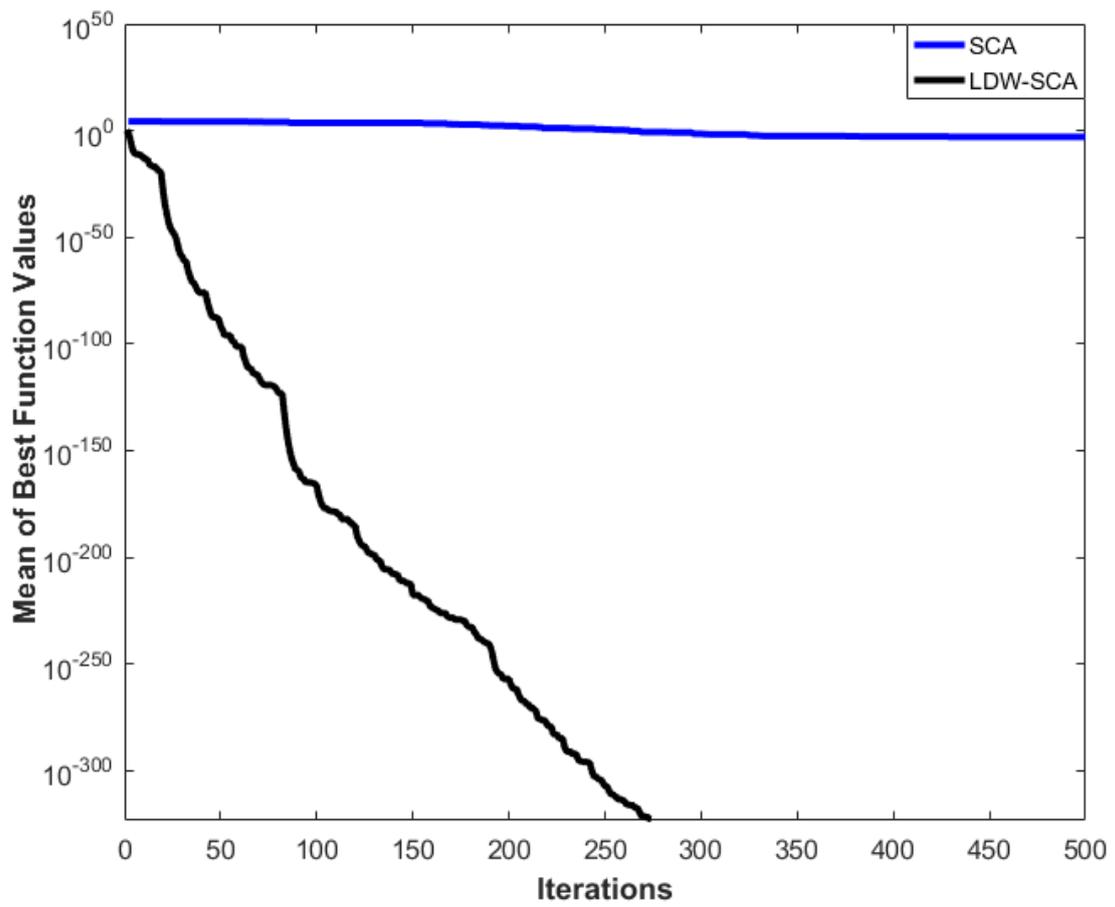

Fig. 4. Comparison of minimization performance of Schwefel 1.2 function ($f_3$) with Dim=30.

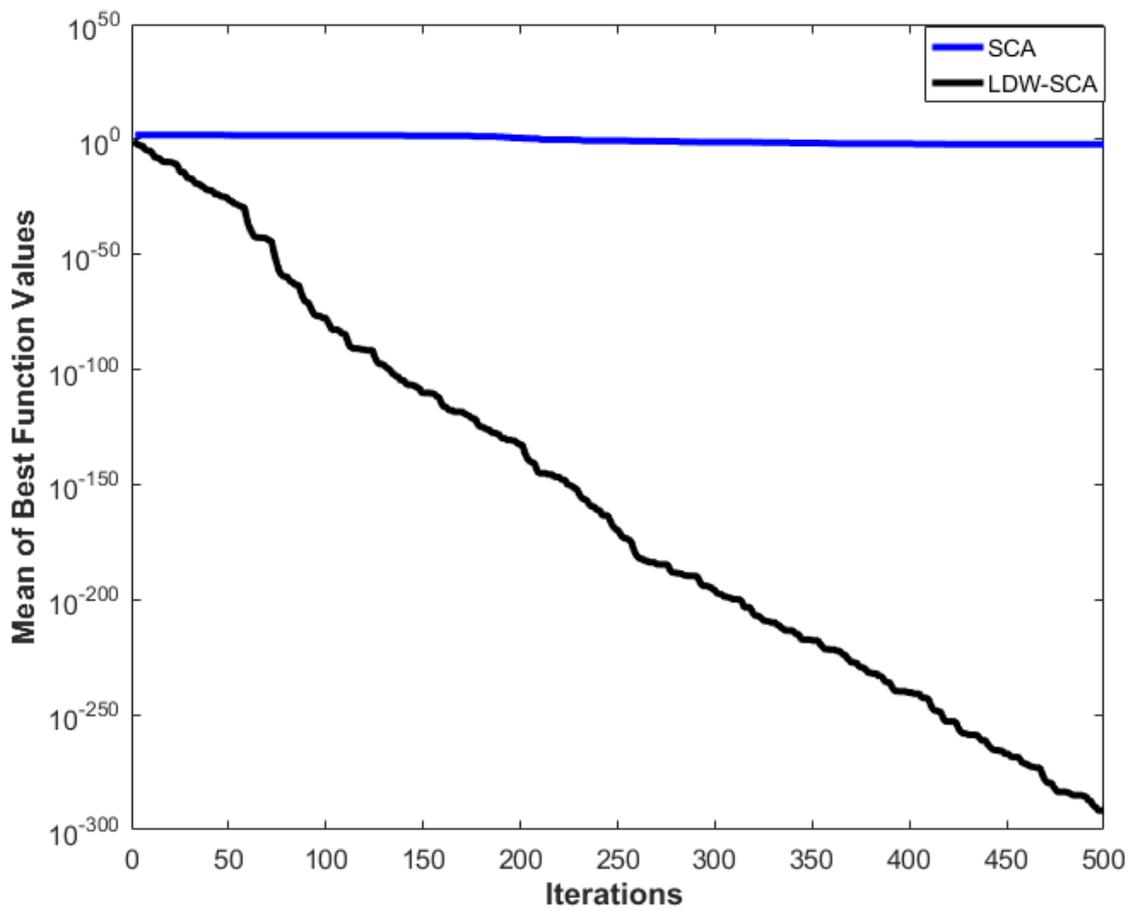

Fig. 5. Comparison of minimization performance of Schwefel 2.21 function ($f_4$) with Dim=30.

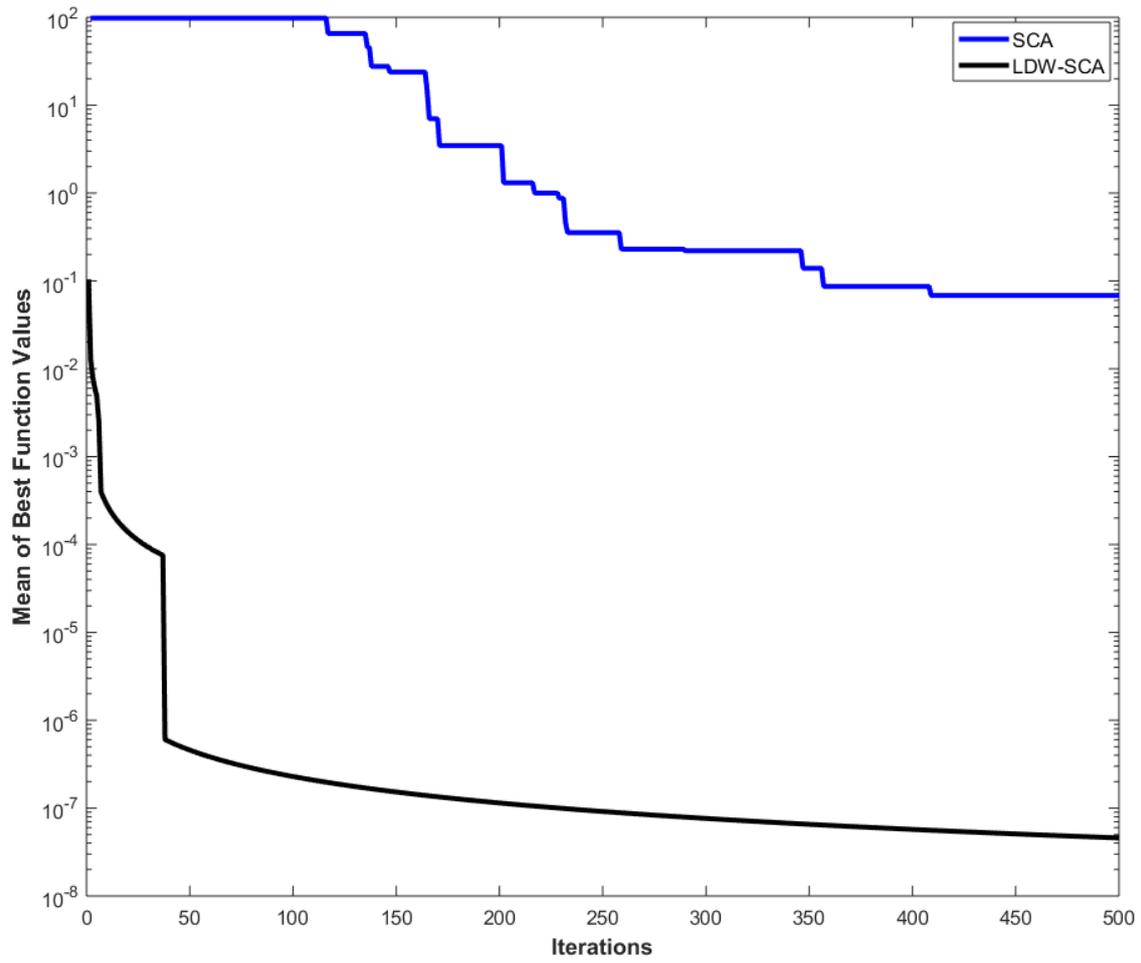

Fig. 6. Comparison of minimization performance of Noise function ($f_7$) with Dim=30.

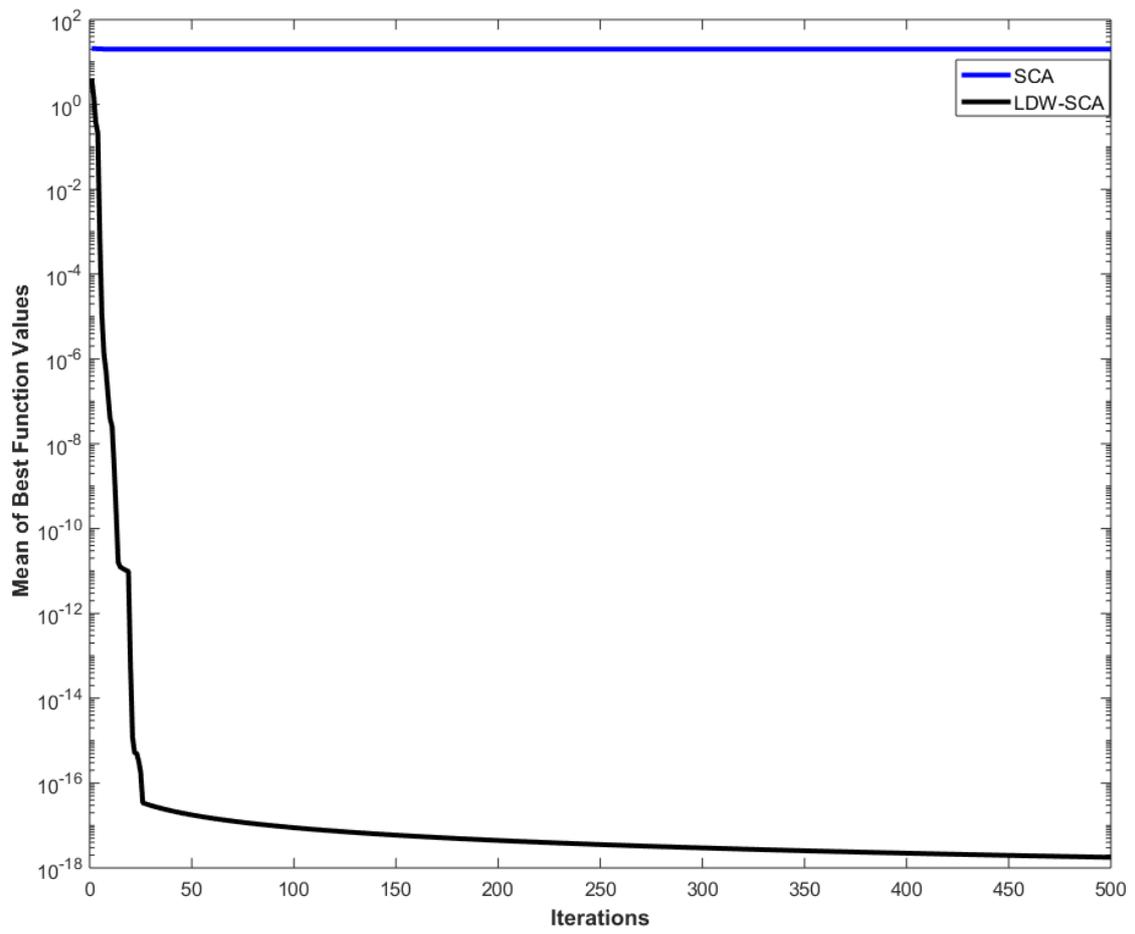

Fig. 7. Comparison of minimization performance of Ackley function ($f_9$) with Dim=30.

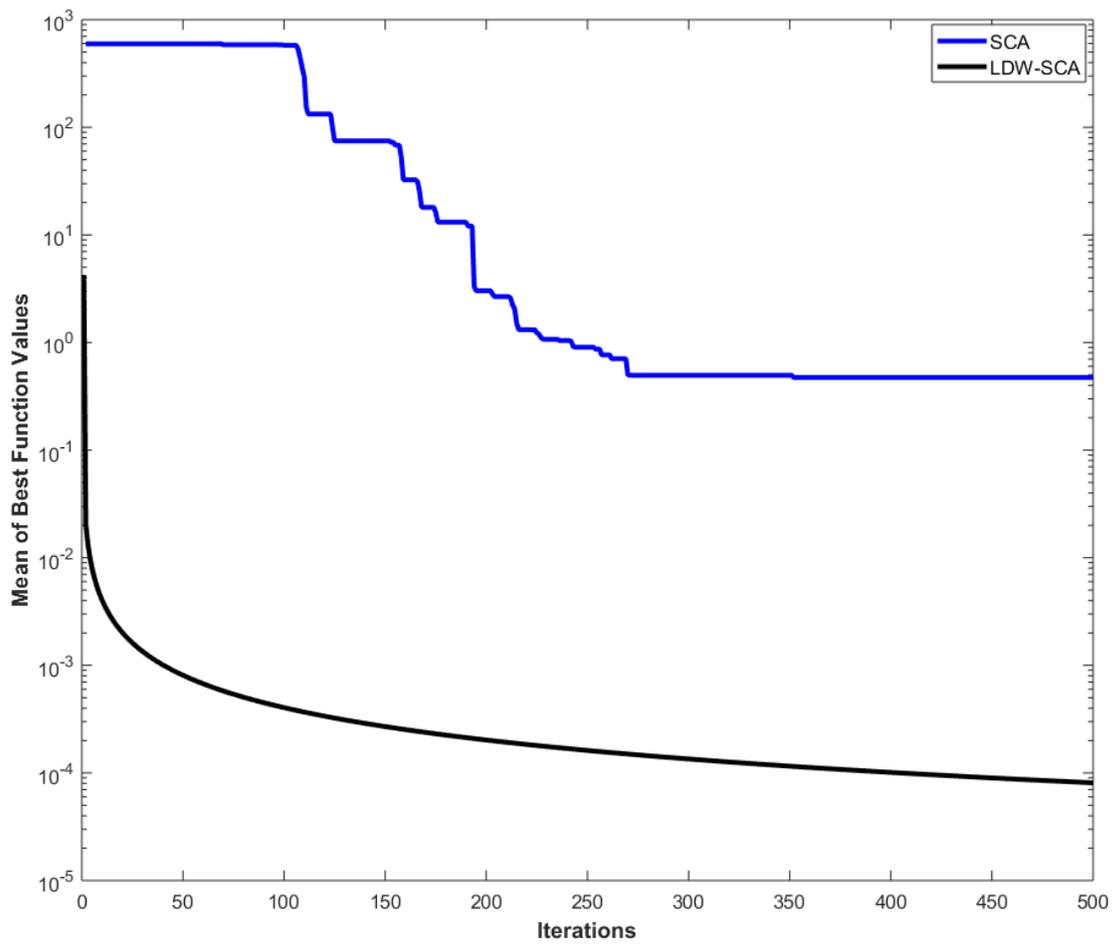

Fig. 8. Comparison of minimization performance of Generalized Penalized 1 function ($f_{11}$) with Dim=30.

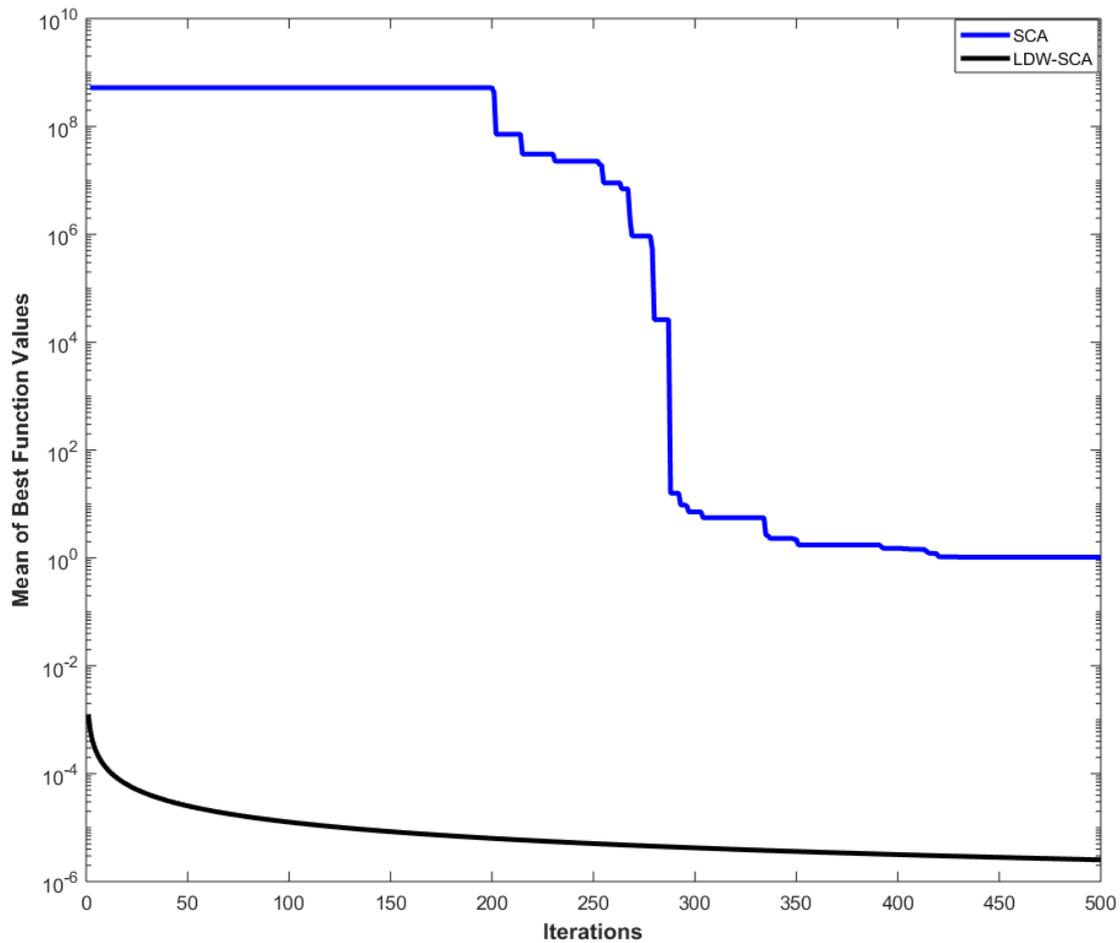

Fig. 9. Comparison of minimization performance of Generalized Penalized 2 function ($f_{12}$) with Dim=30.

Also, LDW-SCSA was compared to vortex search algorithm [23] and PSO2011 [24,25] which is extended version of the PSO and the obtained results were listed in Table 4. In order to compare LDW-SCSA to vortex search and PSO2011 algorithms, 50 particles and 500 iterations were used in LDW-SCSA and 50 particles and 10.000 iterations were used in vortex search and PSO2011 algorithms [23].

Table 4. Comparison results of the LDW-SCA, PSO2011 and vortex search (VS) algorithms (values $< 10^{-16}$ are considered as 0).

| f | Criteria | PSO2011 | VS | LDW-SCA |
|---|----------|---------|----|---------|
|   |          |         |    |         |
| $f_1$ | Mean | 0 | 0 | **0** |
| | S.D. | 0 | 0 | **0** |
| $f_2$ | Mean | 1.0942 | 0 | **0** |
| | S.D. | 0.8707 | 0 | **0** |
| $f_3$ | Mean | 0 | 0 | **0** |
| | S.D. | 0 | 0 | **0** |
| $f_4$ | Mean | - | - | **0** |
| | S.D. | - | - | **0** |
| $f_5$ | Mean | 0.9302 | 0.3678 | **0** |
| | S.D. | 1.7149 | 1.1308 | **0** |
| $f_6$ | Mean | 0.0666 | 0.2 | **2.7767e-04** |
| | S.D. | 0.2537 | 0.4068 | **7.5846e-04** |
| $f_7$ | Mean | 1.6409e-05 | **1.4502e-05** | 2.9754e-05 |
| | S.D. | 5.5658e-05 | **7.3054e-05** | 2.3949e-05 |
| $f_8$ | Mean | 26.1101 | 57.6079 | **0** |
| | S.D. | 5.6866 | 13.9498 | **0** |
| $f_9$ | Mean | 0.6601 | 1.1546e-14 | **8.8818e-16** |
| | S.D. | 0.7114 | 3.6134e-15 | **0** |
| $f_{10}$ | Mean | 0.0047 | 0.0327 | **0** |
| | S.D. | 0.0081 | 0.0185 | **0** |
| $f_{11}$ | Mean | **0.0241** | 0.1146 | 0.1610 |
| | S.D. | **0.0802** | 0.5322 | 0.5874 |
| $f_{12}$ | Mean | **0** | **0** | 0.0030 |
| | S.D. | **0** | **0** | 0.0080 |
| $f_{13}$ | Mean | - | - | **0** |
| | S.D. | - | - | **0** |

## 5. Conclusions and Future Works

In this paper, a novel LDW-SCSA is presented for numerical functions optimization. SCA is modified to achieve better results using dynamic weights by generated logistic map in this work. Steps of the proposed LDW-SCSA are given as follows. Firstly, particles and $r_1$, $r_2$, $r_3$ and $r_4$ are generated randomly. Logistic map is utilized as weight generator. Then, particles are updated using the proposed velocity equation and dynamic weights. After the particles updating, best solution is updated. These steps are repeated until the global optimum or maximum iterations are reached. In order to evaluate performance of this method, the widely used numerical benchmark functions and the widely used optimization algorithms were utilized. The experiments clearly demonstrate that the proposed method achieved better results than others and performance of SCA is increased.

This paper clearly shows that LDW-SCSA will be used for solving real world problems such as feature selection, training neural network, electronics, image processing, etc. In addition, other optimization methods will be improved by using chaotic maps and dynamic weights together.